\pdfoutput=1

\documentclass[11pt]{article}
\usepackage{EMNLP2023}
\usepackage{dblfloatfix}
\usepackage{capt-of}
\usepackage{times}
\usepackage{array}
\usepackage{amsmath}
\usepackage{xcolor}
\usepackage{latexsym}
\usepackage{graphicx}
\usepackage{pifont}
\usepackage{mdframed}
\usepackage{multirow}
\usepackage{hyperref}
\usepackage{tcolorbox} 
\tcbuselibrary{skins} 
\usepackage[T1]{fontenc}
\usepackage{makecell}
\usepackage[utf8]{inputenc}
\usepackage{float}
\usepackage{microtype}
\usepackage{enumitem}

\setlist[enumerate]{topsep=3pt, partopsep=0pt, itemsep=3pt, parsep=3pt}

\usepackage{inconsolata}

\title{Goat: Fine-tuned LLaMA Outperforms GPT-4 on Arithmetic Tasks}

\author{Tiedong Liu \\
  National University of Singapore \\ 
  \texttt{tiedong.liu@u.nus.edu} \\
  \And
  Bryan Kian Hsiang Low \\
  National University of Singapore  \\
  \texttt{lowkh@comp.nus.edu.sg} \\}

\begin{document}
\maketitle
\begin{abstract}
We introduce Goat, a fine-tuned LLaMA model that significantly outperforms GPT-4 on a range of arithmetic tasks. 
Fine-tuned on a synthetically generated dataset, Goat achieves state-of-the-art performance on BIG-bench arithmetic sub-task. In particular, the zero-shot Goat-7B matches or even surpasses the accuracy achieved by the few-shot PaLM-540B. Surprisingly, Goat can achieve near-perfect accuracy on large-number addition and subtraction through supervised fine-tuning only, which is almost impossible with previous pretrained language models, such as Bloom, OPT, GPT-NeoX, etc. We attribute Goat's exceptional performance to LLaMA's consistent tokenization of numbers. To tackle more challenging tasks like large-number multiplication and division, we propose an approach that classifies tasks based on their learnability, and subsequently decomposes unlearnable tasks, such as multi-digit multiplication and division, into a series of learnable tasks by leveraging basic arithmetic principles. We thoroughly examine the performance of our model, offering a comprehensive evaluation of the effectiveness of our proposed decomposition steps. Additionally, Goat-7B can be easily trained using LoRA on a 24GB VRAM GPU, facilitating reproducibility for other researchers. We release our model, dataset, and the Python script for dataset generation.\footnote{\url{https://github.com/liutiedong/goat}.}
\end{abstract}

\section{Introduction}
Large language models (LLMs) have shown remarkable proficiency across a wide range of natural language processing (NLP) tasks \cite{brown2020language, chowdhery2022palm, thoppilan2022lamda}. Notably, GPT-4 \cite{openai2023gpt4} has achieved state-of-the-art performances in such tasks. However, it is surprising that such powerful language models still struggle with elementary arithmetic tasks. The performance of GPT-4 in arithmetic tasks, particularly multiplication and division of large numbers, currently remains far from optimal, with accuracy levels trending toward zero.

\begin{figure}
    \centering
\begin{tcolorbox}[fonttitle=\bfseries,title=Goat-7B, size = normal, label=mybox]
\textbf{3978640188 + 42886272 = }\\
$3978640188 + 42886272 = 4021526460$
\tcbline
\textbf{4523646 minus 67453156}\\
$4523646 - 67453156 = -62929510$
\tcbline
\textbf{Calculate 397 x 4429}
\setlength{\abovedisplayskip}{3pt}
\setlength{\belowdisplayskip}{0pt}
\begin{align*}
& 397 \times 4429  \\
&= 4429 \times (300 + 90 + 7)\\
&= 4429 \times 300 + 4429 \times 90 + 4429 \times 7  \\
&= 1328700 + 398610 + 31003\\
& = 1727310 + 31003 \\
& = 1758313
\end{align*}
\tcbline
\textbf{What is 8914/64?}
\setlength{\abovedisplayskip}{3pt}
\setlength{\belowdisplayskip}{0pt}
\begin{align*}
&8914 - 64 \times 100 = 8914 - 6400 = 2514\\
&2514 - 64 \times 30 = 2514 - 1920 = 594\\
&594 - 64 \times 9 = 594 - 576 = 18\\
&\text{Therefore, } 8914 \div 64 = 139 \text{ R } 18
\end{align*}
\end{tcolorbox}
    \caption{Example of Goat's response on addition, subtraction, multiplication, and division tasks. Prompts are marked in bold, followed by Goat-7B's response.}
    \label{fig:demo}
\end{figure}
In this paper, we present \textbf{Goat}, a fine-tuned language model that is \textbf{GO}od at \textbf{A}rithmetic \textbf{T}asks. Goat achieves state-of-the-art performance in elementary arithmetic, including addition, subtraction, multiplication, and division of integers. We adopt an end-to-end supervised instruction-finetuning paradigm on LLaMA \cite{touvron2023llama}, leveraging a synthetically generated dataset containing around 1 million samples. Unlike previous research on arithmetic computation \cite{lee2023recursion,nogueira2021investigating,nye2021work,qian2022limitations,zhou2022teaching}, our study demonstrates that through supervised fine-tuning alone and without applying any special techniques, our model is capable of generating direct answers for large-number addition and subtraction with near-perfect accuracy in a zero-shot setting. We attribute this exceptional arithmetic ability to LLaMA’s consistent tokenization of numbers and show that this is almost impossible to achieve for previous LLMs such as Bloom \cite{scao2022bloom}, OPT \cite{zhang2022opt}, GPT-NeoX \cite{black-etal-2022-gpt}, Pythia \cite{biderman2023pythia}, etc.

However, the model encounters significant difficulties when generating direct answers for arithmetic tasks like large-number multiplication and division. To overcome this challenge, we propose an approach that categorizes various arithmetic tasks into learnable and unlearnable tasks, subsequently decomposing the unlearnable tasks, such as multi-digit multiplication and division, into a series of learnable tasks by leveraging basic arithmetic principles. Our approach ensures that the intermediate supervision which facilitates the model's learning is also easily understandable and interpretable by humans. We fine-tune our model to generate the proposed CoT before generating the final answer, similar to sketchpad \cite{nye2021work}. Our method outperforms GPT-4's long multiplication and long division methods by a large margin. We assess the performance of our model using BIG-bench \cite{srivastava2022beyond} arithmetic sub-task, and provide a comprehensive evaluation of the effectiveness of our proposed method. Our findings suggest that the model can learn the pattern and generalize to unseen data instead of purely memorizing the computation. Additionally, Goat-7B can be conveniently trained using Low-Rank Adaptation (LoRA) \cite{hu2021lora} technique on a 24GB VRAM GPU, making it easily reproducible for other researchers.

To summarize, our contributions include:
\begin{itemize}    
\item Our model achieves state-of-the-art performance on various elementary arithmetic tasks, including addition, subtraction, multiplication, and division of positive integers (Section~\ref{sec: Experiment}). We show that an open-sourced model fine-tuned on a synthetically generated dataset has the potential to achieve even higher accuracy on arithmetic tasks compared to GPT-4.
\item To the best of our knowledge, we are the first to demonstrate the feasibility that supervised fine-tuning alone can enable LLMs to generate direct answers for certain elementary arithmetic tasks, such as large-number addition and subtraction, without applying any special techniques (Section \ref{sec:addition}). Previously effective chain-of-thought (CoT) methods, such as those used for addition in sketchpad \cite{nye2021work} and LM Tutor \cite{qian2022limitations}, are no longer necessary. The impressive performance is mainly attributed to LLaMA's consistent tokenization of numbers.  
\item To solve large-number multiplication and division, we propose a novel decomposition method based on the learnability of the task, leveraging basic arithmetic principles to ensure human interpretability (Section \ref{sec:Multiplication}).
\item We systematically investigate the proposed decomposition method and demonstrate its effectiveness (Section \ref{sec:Analysis}). We conduct thorough experiments on the decomposition steps in a fully synthetic environment by mitigating many hard-to-control aspects of natural language. Our experimental setup offers an ideal platform to study the impact of CoT and intermediate supervision. 
\item Our end-to-end instruction tuning pipeline can be easily integrated into existing instruction-tuned language models \cite{chiang2023vicuna,alpaca} and potentially enhance their mathematical reasoning for math word problems. We release the model, dataset, and script for generating the dataset. 

\end{itemize}

\section{Related Work}

\subsection{Instruction Tuning}

Instruction tuning \cite{chung2022scaling,ouyang2022training, sanh2022multitask} is a technique used to align pretrained language models with human instructions. It enables targeted customization of LLMs to specific tasks, enhancing their ability to generate more accurate and contextually relevant responses and improving the zero-shot performance. 
The dataset used for instruction tuning can be human-written \cite{ouyang2022training}, machine-generated \cite{peng2023instruction, alpaca, wang2022selfinstruct}, or collected from web \cite{geng2023koala}. Recently, there has been extensive research on fine-tuning LLaMA \cite{touvron2023llama} for various downstream tasks using instruction tuning  \cite{chiang2023vicuna,geng2023koala,alpaca,xu2023baize,li2023chatdoctor}. Creating high-quality instruction tuning datasets can be expensive and time-consuming. In this study, we utilize a simple Python program to generate input-output pairs for arithmetic tasks.

\subsection{Arithmetic Reasoning}

Arithmetic reasoning has been a topic of interest in NLP research for many years \cite{lu2022survey}. Recently, the use of pretrained models \cite{brown2020language, openai2023gpt4} has shown great capabilities in solving math word problems. Particularly, \emph{chain of thought} (CoT) \cite{kojima2022large,wei2022chain,zhou2023least} provides the model with the \emph{intermediate}
steps to derive the final answer. However, studies have shown that LLMs struggle with basic arithmetic computation and often make arithmetic mistakes, even though the reasoning process is correct \cite{cobbe2021training, gao2023pal, schick2023toolformer}. Consequently, one key challenge of arithmetic reasoning, aside from mapping natural language to arithmetic expressions, is how to compute the generated arithmetic expressions with high accuracy.

\begin{table*}[ht!]
\centering
\begin{tabular}{l|lll}
\Xhline{2\arrayrulewidth}
 & \textbf{Task} & \textbf{Input} & \textbf{Output} \\
\hline
\textbf{Learnable}  & Copying & $59265395$ & $59265395$  \\
& Split & $4536$ & $4000 + 500 + 30 + 6$ \\
& Comparison & $8116449, 97863$ & $8116449>97863$ \\
& Ordering & $3568, 9591, 8061$ & $3568, 8061, 9591$ \\
& Addition &  $1270769 + 264985867430$ & $264987138199$ \\
& Subtraction & $40920 - 6173772696$ & $-6173731776$ \\
& Multiplication $nD \times 1D$  & $591714761929184 \times 4$ & $2366859047716736$ \\
& Division $nD \div 1D$ & $339229815457 \div 4$ & $84807453864 \text{ R }  1$\\
\hline
\textbf{Unlearnable} & Multiplication $nD \times mD$ & $6983387 \times 16919$ & $118151924653$  \\
&  Division $nD \div mD$ & $64729486 \div 472$ & $137138 \text{ R } 350$ \\
\Xhline{2\arrayrulewidth}
\end{tabular}
\caption{\label{tab:learn}
Summary and examples of learnable and unlearnable arithmetic tasks. For example, $nD \div 1D$ means $n$-digit by 1-digit division, where $n\geq1$. Unlearnable tasks are mainly multi-digit multiplication and division where $n, m>1$. There are some special cases mentioned in Appendix \ref{sec:special_cases}. 
}
\end{table*}


\subsection{Arithmetic Computation}
Recent studies have explored using external tools to evaluate arithmetic expressions. Toolformer \cite{schick2023toolformer} and GSM8K \cite{cobbe2021training} invoke an external calculator to compute the generated arithmetic expression. PoT \cite{chen2022program} and PAL \cite{gao2023pal} generate programs that can be executed to produce the final answer. While arithmetic can be solved using calculators or programs easily, the ability to perform arithmetic computation is a remarkable trait of human intelligence, and we anticipate LLMs should possess this ability as well. 

Previous studies have evaluated the arithmetic abilities of LLMs. \citet{nogueira2021investigating} have evaluated addition and subtraction tasks. \citet{muffo-etal-2022-evaluating}  have further examined 2-digit multiplication. \citet{yuan2023large} have tested different types of arithmetic operations. CoT seems to be a promising solution for arithmetic computation as well. Similar to humans, autoregressive language model may rely on intermediate supervision to generate the final answer. 
Scratchpad \cite{nye2021work} finetunes the
language models to produce CoT before generating an answer, and has demonstrated effectiveness on 8-digit addition. However, we show that previously effective CoT methods, such as those used for addition in sketchpad \cite{nye2021work} and LM Tutor \cite{qian2022limitations}, are no longer necessary for certain arithmetic tasks like addition. By leveraging simple supervised fine-tuning alone, our model can perform addition and subtraction with sufficiently high accuracy. For challenging tasks like large-number multiplication and division, previous studies \cite{muffo-etal-2022-evaluating,lee2023recursion} either fail to compute or are inefficient. Furthermore, our model is trained end-to-end such that it can follow human instructions. 

\section{Method}
\subsection{Language Model}
LLaMA \cite{touvron2023llama} is a collection of open-source pretrained language models trained on trillions of tokens using publicly available datasets, and achieves state-of-the-art performance on many benchmarks. 

Previous studies \cite{kim-etal-2021-seen, nogueira2021investigating} have shown that tokenization is important for LLM's arithmetic ability. Many commonly-used subword tokenization techniques today are not ideal to represent numbers. However, LLaMA splits each digit into an individual token \cite{yuan2023large}, thereby ensuring \emph{consistent tokenization of numbers}, as shown in  Appendix \ref{sec:token}.

The selection of language models is crucial to our work. We believe the remarkable arithmetic ability demonstrated in this work is mainly attributed to LLaMA's consistent tokenization of numbers. We experimentally verify that other LLMs, such as Bloom, OPT, GPT-NeoX, and Pythia, finetuned on the same arithmetic dataset, cannot match LLaMA's arithmetic ability.

\subsection{Learnability of Arithmetic Tasks}

\citet{wies2023subtask} have provided a theoretical analysis on the use of intermediate supervision for solving composite tasks. Specifically, they have shown that for any family of tasks which on the one hand, are unlearnable, and on the other hand, can be decomposed into a polynomial number of simple sub-tasks, unlearnable composite problems can become learnable by using \emph{intermediate supervision} or \emph{step-by-step CoT}.

Building upon their analysis, we first experimentally categorize learnable and unlearnable tasks. In the context of arithmetic computation, \emph{learnable tasks} generally refer to those for which the model can be successfully trained to generate direct answers, achieving sufficiently high accuracy within a predefined number of training epochs. Conversely, \emph{unlearnable tasks} are those that the model struggles to learn and generate direct answers correctly even with extensive training. While the exact reason behind the varying learnability of tasks is not yet fully understood and requires further investigation, we hypothesize that it is associated with the complexity of the underlying pattern and the size of working memory required for completing the task \cite{bubeck2023sparks}.

We experimentally examine the learnability of these tasks by fine-tuning the model specifically for each task in a simplified synthetic environment (Table \ref{tab:synthetic}). Our recognized learnable and unlearnable tasks are listed in Table \ref{tab:learn}.

The categorization of tasks also aligns with human perception. With practice, humans can mentally calculate the addition and subtraction of two large numbers, writing down the final numerical answer directly from the left (most significant figure) to the right (least significant figure) without the need for sketchpad. However, mentally solving large-number multiplication and division is undeniably a challenging task.

We also observe that our classification of tasks is consistent with the performance of GPT-4. In particular, GPT-4 excels in generating direct answers for large-number addition and subtraction. However, its accuracy significantly drops when it comes to multi-digit multiplication and division tasks. Our observation aligns with the claim made by \citet{bubeck2023sparks} that GPT-4 has a short working memory and performs poorly on composite arithmetic tasks. This is particularly evident in the case of multiplication, which involves multiple steps of addition. The inability of powerful models like GPT-4 to directly solve unlearnable tasks may suggest that generating direct answers for such tasks is extremely challenging, even with extensive training.

It is noteworthy that a task that is learnable for LLaMA may not necessarily be learnable for other LLMs, which is validated in our experiments in Section \ref{sec:otherLLMs}.
Furthermore, not all tasks classified as unlearnable are entirely impossible for the model to learn. For instance, 2-digit by 2-digit multiplication is considered an unlearnable task in our case. However, the model can still learn to generate the direct answer by overfitting to the training set, which contains an exhaustive enumeration of all possible 2-digit multiplication. Nevertheless, the process takes nearly 10 epochs to achieve around 90\% accuracy. In contrast, by inserting our proposed CoT before the final answer, the model can achieve comparable accuracy in 2-digit multiplication with only 1 epoch of training. These findings align with the claim \cite{wies2023subtask} that the presence of intermediate supervision facilitates the learning process.

\subsection{Addition and Subtraction}
\label{sec:addition}
Addition and subtraction tasks are learnable, as with supervised fine-tuning alone, the model exhibits a remarkable ability to accurately generate direct numerical answers. The model successfully captures the underlying patterns of the arithmetic operations. This is evident from the model's near-perfect accuracy on the unseen test set, despite being trained on a very limited subset of the data. It is worth mentioning that addition and subtraction operations do not require the use of CoT. This contrasts with previous studies that have employed CoT for addition and subtraction tasks \cite{lee2023recursion,nye2021work,qian2022limitations}.

\subsection{Multiplication}
\label{sec:Multiplication}
We experimentally verify that $n$-digit by 1-digit multiplication is learnable. In contrast, multi-digit multiplication poses significant challenges for the model, suggesting it to be an unlearnable task. To overcome this issue, we adopt a similar strategy used in sketchpad \cite{nye2021work}, which fine-tunes the LLMs to generate CoT before generating the answer. Specifically, we propose a CoT that decomposes the multi-digit multiplication into a series of 5 learnable sub-tasks: (1) \textbf{extraction}: extract the arithmetic expression from the natural language instruction, 
(2) \textbf{split}: split the smaller number of the two into place values, 
(3) \textbf{expansion}: expand the sum based on the distributive property, 
(4) \textbf{product}: compute each product simultaneously, and  
(5) \textbf{adding term by term}: add the first two terms and copy the rest, and the final sum is obtained.


Consider the example in Fig.~\ref{fig:demo}. Firstly, the arithmetic expression $397 \times 4429$ is extracted from the instruction, which can be considered as a ``copying'' task. Secondly, $397 \times 4429 = 4429 \times (300 + 90 + 7)$ involves two learnable tasks. The larger number of the two is placed in front and then the smaller one is split, which is similar to ``ordering'' and ``split'' learnable tasks. The ordering ensures that there are fewer summation terms in the next step, thereby reducing the CoT length. Thirdly, the sum is expanded using distributive law: $4429 \times (300 + 90 + 7) = 4429 \times 300 + 4429 \times 90 + 4429 \times 7$, which is similar to ``copying'' task. Next, $ 4429 \times 300 + 4429 \times 90 + 4429 \times 7 = 1328700 + 398610 + 31003$ where the products are computed at once by applying ``multiplication $n$-digit by 1-digit'' with zeros copied at the end of each product. Finally, we take the sum of the first two terms at each step, and copy the rest terms, leveraging ``addition'' and ``copying''. Hence, a composite unlearnable task is broken down into simpler tasks that are all learnable.

\begin{table*}[t!]
\centering
\begin{tabular}{lclcl}
\Xhline{2\arrayrulewidth}
\textbf{Task} & \textbf{Learnable} & \textbf{Prompt} & \textbf{CoT} & \textbf{Target}\\
\hline
\textbf{ADD} & \ding{51} & 1463456 + 2107 &\ding{55} & 1463456 + 2107 = 1465563 \\
\textbf{SUB} & \ding{51} & 2348233 minus 483579? &\ding{55} & 2348233 - 483579 = 1864654 \\
\textbf{MUL} \\
\hspace{2ex}$nD \times 1D$ & \ding{51}& 593295 times 7 &\ding{55} &  593295 * 7 = 4153065\\
\hspace{2ex}$nD \times mD$ & \ding{55}& Calculate 24 x 79 &\ding{51} & 24 * 79 = 24 * (70 + 9) = 24 * 70 + \textbackslash\\ &&&&  24 * 9 = 1680 + 216 = 1896 \\
\textbf{DIV}\\
\hspace{2ex}$nD \div 1D$ & \ding{51}& Please tell 3651803/7 &\ding{55} & 3651803 / 7 = 521686 R 1\\
\hspace{2ex}$nD \div mD$ & \ding{55}  & What is 2546/38? &\ding{51}  &
2546 - 38 * 60 = 2546 - 2280 = 266 \\ &&&&
266 - 38 * 7 = 266 - 266 = 0 \\ &&&&
Therefore, 2546 / 38 = 67 \\
\Xhline{2\arrayrulewidth}
\end{tabular}
\caption{\label{tab:CoT}
Examples of prompts and targets for fine-tuning LLaMA. ``\textbackslash nAnswer: '' is appended at the end of each prompt. It should be noted that there are a few special cases when CoT is not required (see Appendix \ref{sec:special_cases}). 
}
\end{table*}

\subsection{Division}
Similarly, we observe that $n$-digit by 1-digit division is learnable. However, multi-digit division is unlearnable. We design a novel CoT leveraging a modified slow division method based on the following recurrence equation
$$R_j - D \times (q_{n-(j+1)} \times 10^j) = R_{j+1} $$
where $R_j$ is the $j$-th partial remainder of the division,
$q_{n-(j+1)}$ is the digit of the quotient in position $n-(j+1)$ numbered from least significant 0 to most significant $n-1$,
$n$ is the number of digits in the quotient, and
$D$ is the divisor. Specifically, the main idea is to subtract multiples of the divisor from the dividend until the remainder is less than the divisor.


Here is a detailed breakdown of the CoT used in Fig.~\ref{fig:demo}. Consider the first iteration (first equation). The first step $8914 - 64 \times 100$ requires the model to copy the dividend and the divisor, and subsequently generate a number $q_{n-(j+1)}\times10^j$ such that the product of $q_{n-(j+1)}\times10^j$ and the divisor $D$ is less than or equal to the partial remainder $R_j$. This inherently involves two learnable tasks: ``$n$-digit by 1-digit multiplication'' and ``comparison''. We experimentally show that this composite task is learnable. The second step $8914 - 64 \times 100 = 8914 - 6400$ mainly involves a ``copying'' task and an ``$n$-digit by 1-digit multiplication'' task. The third step $8914 - 6400 = 2514$ leverages ``subtraction''. The process iterates until the leftover is less than the divisor, which implies the model has to implicitly learn comparison. Finally, the model generates the quotient by combining all $q_{n-(j+1)}$'s in previous iterations, which can be considered as the inverse of the ``split'' task, and finally copies the remainder if it is not zero.

A summary of prompts and expected output for various tasks are shown in Table \ref{tab:CoT}.

\subsection{Settings} 
\label{sec:settings}
In this paper, we consider the addition and subtraction of two positive integers with each containing up to 16 digits. It is worth noting that the result of subtraction can be negative. To limit the maximum generated sequence length, we consider the multiplication of two positive integers whose product falls within 12 digits, and the division of two positive integers resulting in a quotient within 6 digits where the dividend is less than 12 digits. Since we focus on arithmetic tasks of integers, we aim to obtain the least positive remainder in the case when it is not divisible.

In Section \ref{sec:extrapolation}, we present an analysis showcasing the limited extrapolation capabilities of fine-tuned LLMs. Consequently, input data that falls outside the distribution of the training data is unlikely to yield reasonable answers. Our method potentially applies to numbers with more digits, though the training cost will increase correspondingly.

\subsection{Dataset}
We generate the dataset synthetically using a Python script. The dataset consists of around 1 million question-answer pairs. The answer contains the proposed CoT as well as the final numerical output. The numbers are randomly generated, hence ensuring a very low probability of instances being duplicated, although small numbers may be sampled multiple times. We sample from log space to ensure the numbers are equally likely to be sampled from different orders of magnitude, which is similar to the sampling method used by \citet{lee2023recursion}. The details of the dataset are presented in Appendix \ref{sec:dataset}.

\subsection{Fine-tuning}
To enable the model to solve arithmetic problems based on instructions and facilitate natural language question answering, we generate hundreds of instruction templates using ChatGPT (Table \ref{tab:template}). During the instruction tuning process, we randomly select a template for each arithmetic input from the training set, and fine-tune LLaMA-7B similar to the method used in Alpaca \cite{alpaca}. We apply various techniques to enhance the model's adaptability to diverse question formats, such as randomly removing spaces between numbers and symbols in the arithmetic expression, replacing ``*'' with ``x'' or ``times'', etc.

Goat-7B can be easily fine-tuned using LoRA on a 24GB VRAM GPU. In particular, the fine-tuning process for a specific arithmetic sub-task, such as 8-digit addition using 100K instances, takes only approximately 1.5 hours on an A10 GPU to achieve near-perfect accuracy. The training hyperparameters are listed in Appendix~\ref{crap}.

\begin{table*}
\centering

\renewcommand{\arraystretch}{1.2}
\begin{tabular}{@{}c|@{ }ccccc@{ }|@{ }ccc@{}}
\Xhline{3\arrayrulewidth}
\textbf{Task} & \multicolumn{5}{c}{\textbf{BIG-bench}} & \multicolumn{3}{c}{\textbf{Extra Tasks}} \\
\Xhline{3\arrayrulewidth}
\textbf{ADD} & 1D & 2D & 3D & 4D & 5D & 8D$+$8D & 16D$+$8D & 16D$+$16D \\
\hline
GPT-4 & 100/100 & 100/100 & 99.6/99.9 & 98.8/99.6 & 94.1/98.5 & 92.1/98.3 & 9.4/70.4 & 94.1/99.5 \\
Goat-7B & 100/100 & 100/100 & 99.4/99.8 & 98.3/99.5 & 98.1/99.4  & 97.8/99.4 & 97.1/99.6 & 97.6/99.7 \\
\hline
\textbf{SUB} & 1D & 2D & 3D & 4D & 5D & 8D$-$8D & 16D$-$8D & 16D$-$16D \\
\hline
GPT-4 & 100/100 & 100/100 & 99.2/99.6 & 98.9/99.6 & 92.4/98.1 & 70.5/91.5 & 10.6/68.8 & 59.6/88.2 \\
Goat-7B & 100/100 & 100/100 & 99.7/99.9 & 98.6/99.6 & 98.4/99.5 & 96.8/99.3 & 95.8/99.2 & 96.3/99.3 \\
\hline
\textbf{MUL} & 1D & 2D & 3D & 4D & 5D & 1D$\times$16D & 4D$\times$8D & 6D$\times$6D \\
\hline
GPT-4 & 100/100 & 99.4/99.8 & 30.3/83.0 & 5.3/61.8 & 0.0/47.9 & 61.5/92.3 & 0.0/45.9 & 0.0/49.8 \\
Goat-7B & 100/100 & 100/100 & 97.8/99.4 & 96.9/99.2 & 96.7/99.3 & 99.7/99.9 & 88.1/97.8 & 96.8/99.5 \\
\hline
\textbf{DIV} & 1D & 2D & 3D & 4D & 5D & 16D$\div$1D & 6D$\div$3D & 12D$\div$6D \\
\hline
GPT-4 & 100/100 & 100/100 & 94.5/96.3 & 90.9/92.1 & 53.4/73.2 & 54.0/84.3 & 6.4/48.6 & 0.0/29.5 \\
Goat-7B & 100/100 & 100/100 & 99.5/99.7 & 99.0/99.5 & 96.5/98.1 & 99.0/99.7 & 94.1/96.1 & 89.3/93.5 \\
\Xhline{3\arrayrulewidth}
\end{tabular}
\caption{The result of GPT-4 and Goat-7B on BIG-bench Arithmetic sub-task and extra selected arithmetic tasks, using metrics Exact String Match/Digit Match (Appendix \ref{sec:metric}), shown in percentage. We test GPT-4 and Goat with exactly the same questions and prompts. We evaluate GPT-4 using the API version on May 10th. For Big-bench tasks, $nD$ refers the $n$-digit by $n$-digit operation, except for division where $nD$ means $n$-digit by $m$-digit where $m\leq n$. BIG-bench only includes division operation without remainder, whereas in extra tasks we include the cases where the remainder is not zero and ask GPT-4 to output the answer in "quotient R remainder" format. It should be noted that we exclude the BIG-bench test data from our training dataset as much as possible, although the overlap is unavoidable for operations involving small numbers.
}
\label{tab:result}
\end{table*}

\section{Experiments}
\label{sec: Experiment}

We evaluate our model using BIG-bench arithmetic dataset \cite{srivastava2022beyond}, as well as our extra selected tasks. The results are shown in Table \ref{tab:result}. Notably, in a zero-shot setting, Goat-7B achieves comparable or even higher accuracy on BIG-bench compared to the few-shot PaLM-540B.

\subsection{Metric}
We first compute the accuracy based on the standard exact string match (Appendix \ref{sec:metric}). We observe that GPT-4's accuracy under exact string match is almost identically zero on tasks involving large numbers. However, in many cases where the final answer is incorrect, the majority of digits in the generated answer align with the target number, with only a few digits being incorrect. Inspired by recent study on the emergent abilities of LLMs \cite{schaeffer2023emergent}, we include a digit match metric that can reflect the per-token error rate of the output, as each digit is uniquely represented by a token in LLaMA.

\subsection{Comparison}
Comparing the performance of Goat and GPT-4 for large-number multiplication and division may seem unfair, as GPT-4 generates direct answers while Goat relies on CoT. Hence, we also evaluate GPT-4's performance with CoT by appending ``Solve it step by step'' at the end of each prompt. By default, GPT-4 uses long multiplication and long division methods. However, we observe that generating CoT only leads to marginal improvement in accuracy. In some cases, the intermediate steps from long multiplication and division are incorrect, but surprisingly the final answer is correct. This implies that GPT-4 does not effectively take advantage of intermediate supervision from CoT to improve the final output. 
We identify the following 3 common errors from GPT-4's solution, which results in incorrect final answers: (1) the alignment of corresponding digits, (2) copying of numbers, and (3) the intermediate result from $n$-digit by 1-digit multiplication.

Additionally, we observe that GPT-4 performs reasonably well on $8D + 8D$ and $16D + 16D$ tasks, but fails on most $16D + 8D$ tasks, though intuitively $16D + 8D$ should be relatively easier than $16D + 16D$. While the exact reason for this remains unclear, one possible factor could be GPT-4's inconsistent number tokenization (Table \ref{tab:tokenization}), which makes it difficult to align the corresponding digits of two numbers.

\section{Analysis}
\label{sec:Analysis}

\subsection{Ablation study}
\begin{figure}[h]
  \centering
  \includegraphics[width=0.5\textwidth]{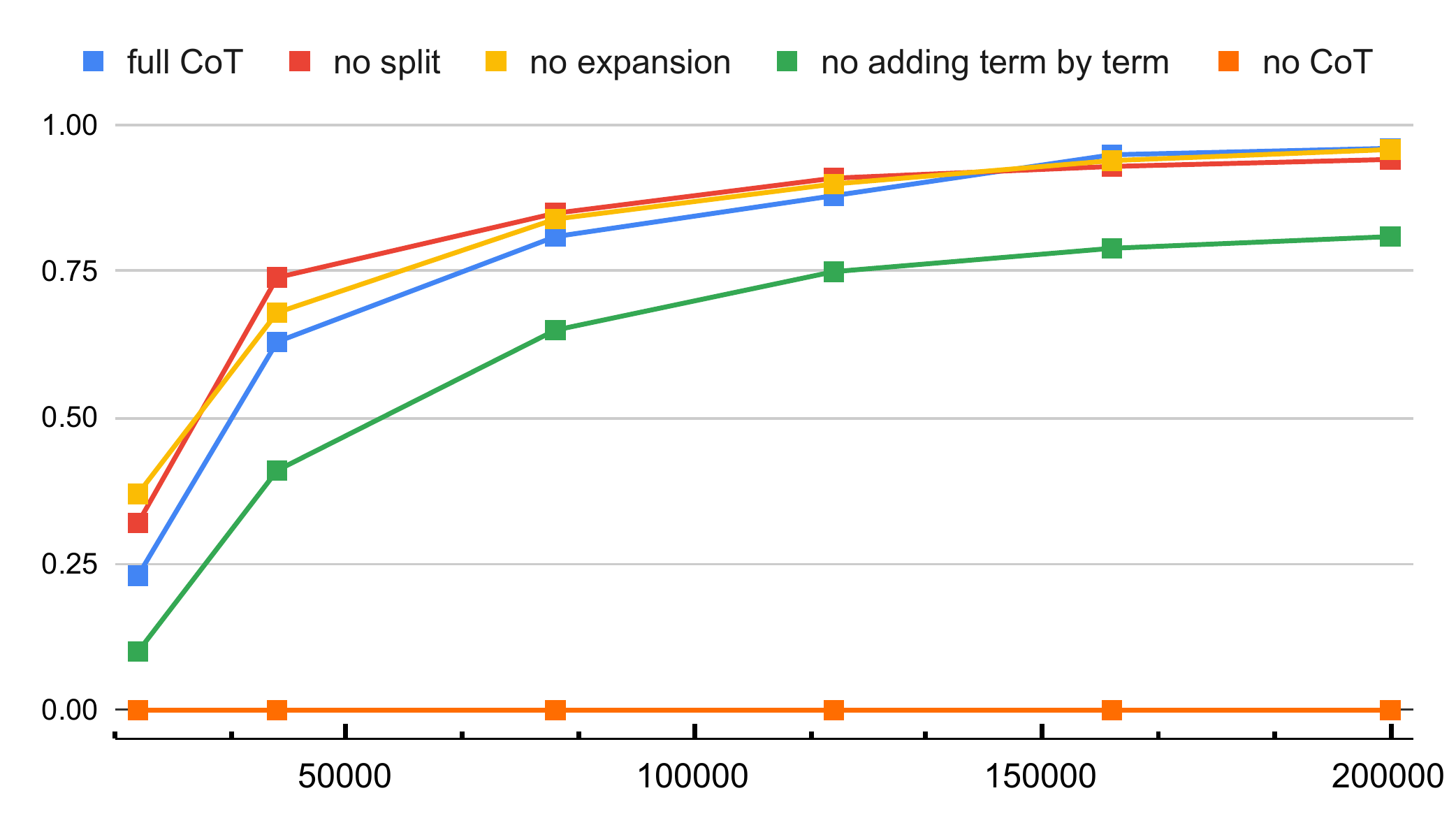}
  \caption{Accuracy (exact string match) against the number of samples seen during the training of $4D \times 4D$ task. Evaluated on the same randomly generated unseen test set using training checkpoints.}
  \label{fig:multiplication}
\end{figure}

\begin{figure}[h]
  \centering
  \includegraphics[width=0.5\textwidth]{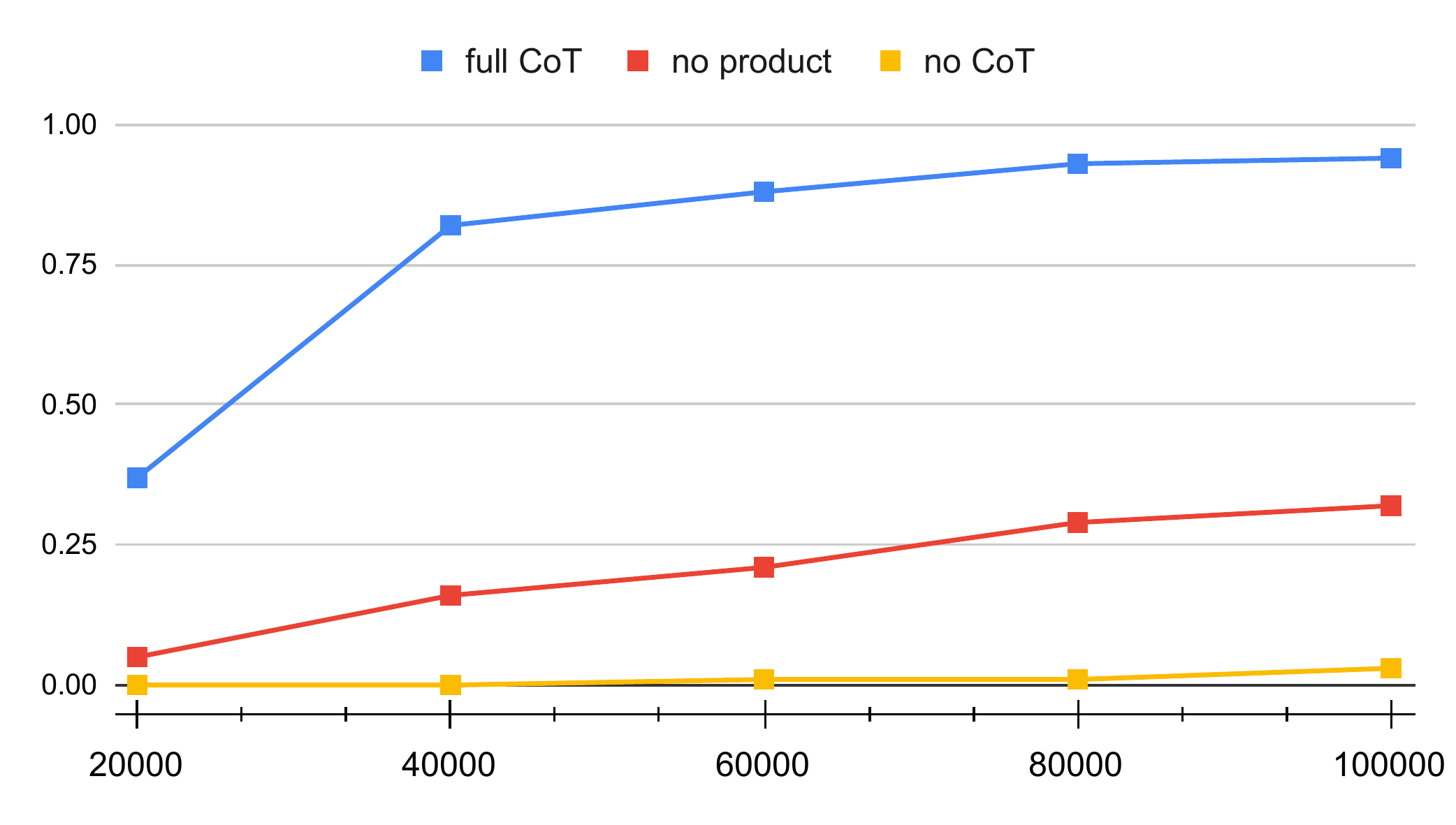}
  \caption{Accuracy (exact string match) against the number of samples seen during the training of $6D \div 3D$ task. Evaluated on the same randomly generated unseen test set using training checkpoints.}
  \label{fig:division}
\end{figure}
Here we want to study the usefulness and effectiveness of each intermediate decomposition step. Specifically, for multiplication (Fig.~\ref{fig:multiplication}), we compare the accuracy of 4-digit by 4-digit multiplication by removing one particular step in the CoT, including split, expansion, adding term by term (referring to \ref{sec:ablation_appendix}), as well as no CoT. For division (Fig.~\ref{fig:division}), we compare the accuracy of 6-digit by 3-digit division after removing the middle step that computes the product (referring to \ref{sec:ablation_appendix}), as well as no CoT. To minimize the impact caused by natural language, we conduct an ablation study in a simplified synthetic environment (Table \ref{tab:synthetic}).

The multiplication results suggest that the ``adding term by term'' step plays a crucial role in obtaining the final answer. In contrast, the ``split'' and ``expand'' steps have minimal impact, and can potentially be omitted for generating more concise CoT. This can be attributed to the nature of these two intermediate steps, which primarily involve simple and learnable tasks like copying and comparison. Nevertheless, we still retain these steps to ensure human interpretability.

The accuracy of exact string match without CoT remains consistently at zero for both $4D\times 4D$ multiplication and $6D\div 3D$ division. This further showcases the validity of our approach, as breaking down complex arithmetic tasks into a series of learnable tasks can indeed facilitate the training process for LLMs.

\subsection{Extrapolation}
\label{sec:extrapolation}
Extrapolation refers to the ability of the model to predict data that lies out-of-distribution (OOD) of training data. We test addition for numbers larger than those in the training data distribution. The results reveal that the model has limited extrapolation capabilities. There is a gradual drop in accuracy, as the test set deviates further from the training set. This observation is consistent with the result reported in \cite{kim-etal-2021-seen}, highlighting a limitation of our fine-tuned model and underscoring the significance of training data distribution.
\begin{figure}[h]
  \centering
  \includegraphics[width=0.5\textwidth]{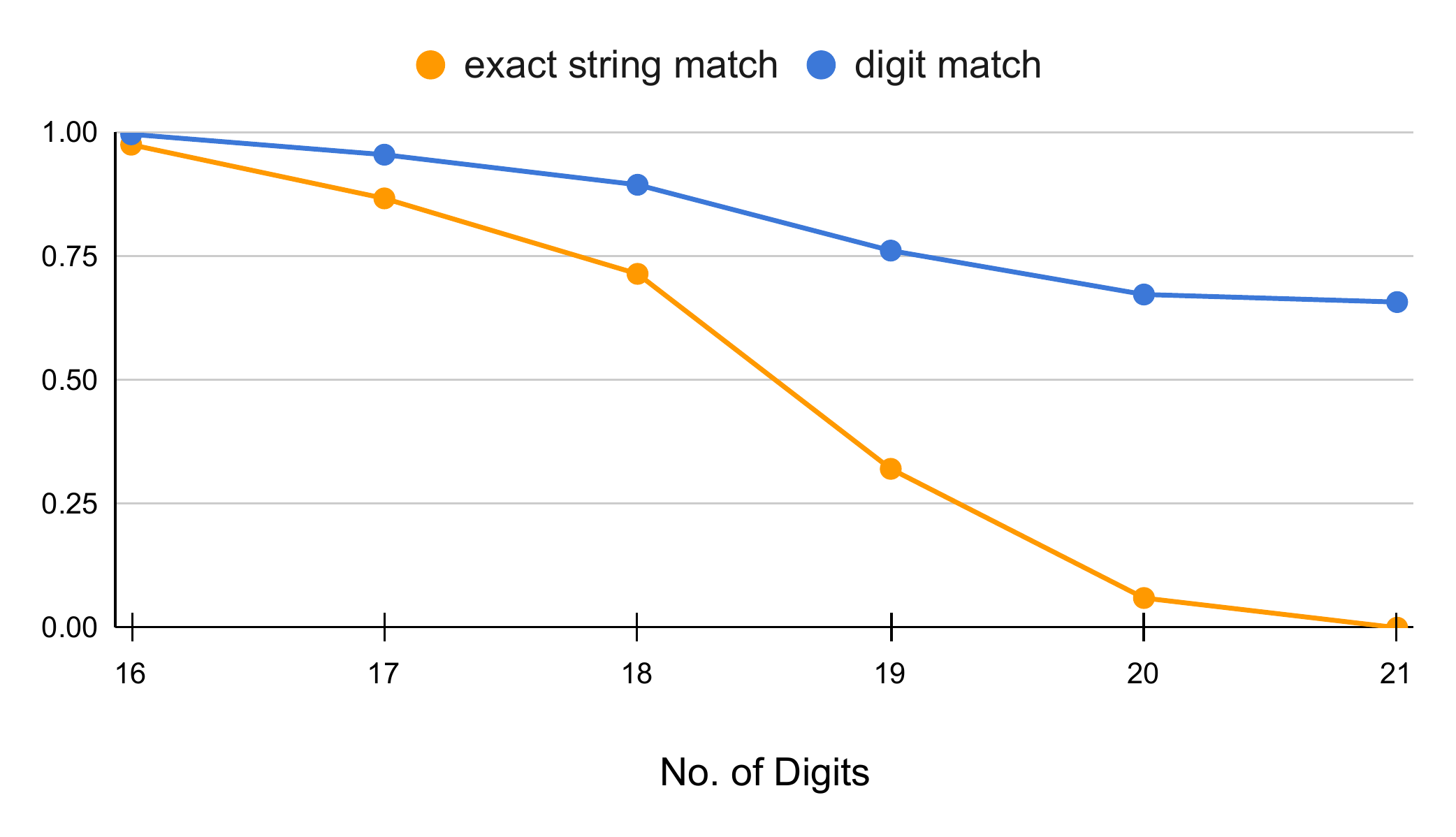}
  \caption{Accuracy against the number of digits for the addition task. The model is trained up to 16D+16D, and tested on 17D+17D onward.}
  \label{fig:extrapolation}
\end{figure}


\subsection{Comparison with Other LLMs}
\label{sec:otherLLMs}
We conduct comprehensive experiments on a variety of LLMs, including Bloom, OPT, GPT-J, GPT-NeoX, and Pythia. These models are fine-tuned using the identical dataset as that for Goat, maintaining consistency in the training hyperparameters. Our experiment shows that they all struggle with arithmetic tasks. Even for tasks that are considered learnable for LLaMA, such as multi-digit addition, the loss during fine-tuning is significantly higher than that of LLaMA. The observation underscores the claim made in \cite{nogueira2021investigating} that tokenization is a crucial factor in the performance of arithmetic tasks.





\subsection{Few-shot Prompting with GPT-4}
GPT-4 demonstrates powerful in-context learning abilities. We further examine the effectiveness of our proposed decomposition method for solving large-number multiplication and division by using few-shot prompting with GPT-4 (see Appendix \ref{sec:fewshot}).
We observe that our decomposition method allows GPT-4 to generate correct answers more frequently than using its default long multiplication and division methods. This further supports the effectiveness and validity of our approach. Examples of the prompt and output are shown in Appendix \ref{sec:fewshot}. 

\section{Limitations}
Humans are capable of performing multiplication and division on arbitrarily large numbers, providing sufficient time and space for calculations. In contrast, LLMs often suffer from extrapolation problems. The models are unlikely to generate reasonable answers if the input deviates significantly from the distribution of training data. To enhance the human interpretability of intermediate supervision, we use the straightforward CoT that follows simple basic arithmetic rules. However, this design may not be the most efficient way to facilitate the final answer generation. There are potentially more suitable multiplication and division algorithms for the model to learn.
Besides, our research only focuses on elementary arithmetic operations involving integers. Nevertheless, we anticipate that our method could be applicable to decimal computation as well. 

\section{Conclusion}
In summary, we demonstrate the feasibility that supervised fine-tuning alone can enable LLMs to perform certain basic arithmetic operations with high accuracy. With our proposed CoT, our model achieves state-of-the-art performance on various elementary arithmetic tasks. Our research offers an excellent platform for investigating the mechanism of working memory and the influence of intermediate supervision on text generation. Our method can be easily integrated with other instruction-tuned LLMs and has the potential to further enhance arithmetic reasoning abilities in solving math word problems.


\bibliography{anthology,custom}
\bibliographystyle{acl_natbib}

\newpage
\appendix

\section{Hyperparameters}
\label{crap}

\begin{table}[htbp]
  \centering

  \begin{tabular}{cc}
\Xhline{3\arrayrulewidth}
    Hyperparameter & Value  \\
    \hline
    batch size   & 128    \\
    learning rate   & 0.0003   \\
    lora r   & 64    \\
    lora alpha & 64 \\
    lora target module & q, v, k, o\\
    lora dropout & 0.05 \\
    epoch & 1\\
\Xhline{3\arrayrulewidth}
  \end{tabular}
\caption{Hyperparameters for fine-tuning LLaMA-7B.}
\label{tab:training}
\end{table}


\section{Tokenization}
\label{sec:token}
\citet{nogueira2021investigating} demonstrate that models with inconsistent tokenization of numbers barely learn the addition of 2-digit numbers, and it completely fails to learn the addition of larger numbers. Specifically, it has an accuracy of zero for 5 digits or more. They attribute this failure to the lack of systematic tokenization of individual digits. For instance, “123” might be tokenized as “12” and “3”, while “234” might be tokenized as “2” and “34”. Consequently, the model is required to learn that the embedding of a token may represent either a single digit or two digits and so on. Hence, it might be challenging for the model to learn to map an embedding to a number when the number of digits it represents changes irregularly. In Table \ref{tab:tokenization}, we compare number tokenization across different LLMs.


\begin{table*}
\centering
\begin{tabular}{lll}
\Xhline{2\arrayrulewidth}
\textbf{Model}& \textbf{Number} & \textbf{Tokenization} \\
\hline
LLaMA& 
74815 & [1, 29871, 29955, 29946, 29947, 29896, 29945] \\ 
&7481 & [1, 29871, 29955, 29946, 29947, 29896]\\ 
&748 & [1, 29871, 29955, 29946, 29947]\\ 
&74 & [1, 29871, 29955, 29946]\\ 
&7 & [1, 29871, 29955]\\ 
\hline
GPT-4 & 
74815 & [20338, 868] \\ 
&7481 & [20338, 16]\\ 
&748 & [20338]\\ 
&74 & [5728]\\ 
&7 & [22]\\ 
\hline
Bloom& 
74815 & [88241, 2057] \\ 
&7481 & [88241, 20]\\ 
&748 & [88241]\\ 
&74 & [8771]\\ 
&7 & [26]\\ 
\hline
OPT& 
74815 & [2, 39373, 996] \\ 
&7481 & [2, 406, 34490]\\ 
&748 & [2, 39373]\\ 
&74 & [2, 5243]\\ 
&7 & [2, 406]\\ 
\hline
Pythia& 
74815 & [24, 2385, 1010] \\ 
GPT-NeoX-20B&7481 & [24, 34474]\\ 
MPT-7B &748 & [24, 2385]\\ 
&74 & [3566]\\ 
&7 & [24]\\ 
\hline
GPT-J& 
74815 & [48246, 1314]\\ 
GPT-Neo &7481 & [22, 40271]\\ 
&748 & [48246]\\ 
&74 & [4524]\\ 
&7 & [22]\\ 
\hline
ChatGLM& 
74815 & [5, 25, 16, 23, 9, 15, 130001, 130004]\\ 
&7481 & [5, 25, 16, 23, 9, 130001, 130004]\\ 
&748 & [5, 25, 16, 23, 130001, 130004]\\ 
&74 & [5, 25, 16, 130001, 130004]\\ 
&7 & [5, 25, 130001, 130004]\\ 
\Xhline{2\arrayrulewidth}
\end{tabular}
\caption{Comparison of number tokenization of various LLMs. It should be noted that ChatGLM also splits each digit into an individual token. Evaluating ChatGLM's arithmetic abilities will be left as future work.}
\label{tab:tokenization}
\end{table*}

\begin{table*}[h]
\centering
\begin{tabular}{cl}
\Xhline{2\arrayrulewidth}
\textbf{Index} & \textbf{Template}\\
\hline
1 & \{arithmetic\} = \\ 
2 & What is \{arithmetic\}?\\ 
3 & Compute \{arithmetic\}\\ 
4 & Solve \{arithmetic\}\\ 
5 & Determine \{arithmetic\}\\ 
6 & Find \{arithmetic\}\\ 
7 & What is the result of \{arithmetic\}?\\ 
8 & Please help me calculate \{arithmetic\}.\\ 
9 & Solve the following problem: \{arithmetic\}\\ 
10 & I am looking for the value of \{arithmetic\}. Can you help?\\ 
11 & What is the numerical value of \{arithmetic\}?\\ 
12 & Help me obtain \{arithmetic\}\\ 
13 & Show me the result of \{arithmetic\}?\\ 
14 & Kindly calculate \{arithmetic\} for me.\\ 
15 & Determine the value for \{arithmetic\}.\\ 
16 & Can you please compute \{arithmetic\}?\\ 

17 & Find the numerical value of  \{arithmetic\}?\\ 
18 & I would appreciate it if you could assist me in calculating  \{arithmetic\}.\\ 
19 & Please work out  \{arithmetic\}.\\ 
20 & What is the answer to  \{arithmetic\}?\\ 

\ldots & \ldots \\ 
\Xhline{2\arrayrulewidth}
\end{tabular}
\caption{Example templates to fine-tune arithmetic tasks with natural language instructions, generated by ChatGPT. During training, \{arithmetic\} is replaced by the randomly generated arithmetic expression, like $3425 * 5823$.}
\label{tab:template}
\end{table*}

\section{Metric}
\label{sec:metric}
Exact string match is defined as 1 if the output string exactly matches the target string, and 0 otherwise. Then we take the average of exact string match for each task. Char error rate (CER) is defined as the percentage of characters that were incorrectly predicted. We compute CER using Python torchmetrics package. Then we define digit match accuracy as $1 - cer$. We include this metric because, for difficult tasks, the exact string match could be identically zero, making it hard to evaluate the performance. In many cases, both GPT-4 and Goat may have very few incorrect digits in the middle of the generated answer, and the number of digits in the generated answer generally matches the target number.

\section{Simplified Synthetic Environment}
We use the simplified synthetic environment to study the effectiveness of various CoT, by avoiding many hard-to-control aspects of natural languages. The difference between this and Goat is that we use a more structured prompt without any instruction template and a straightforward completion of the task. This enables easy comparison between the model's performance on different tasks, allowing us to examine the learnability of various sub-tasks and explore the effectiveness of the proposed CoT. The input and output examples for the simplified synthetic environment are shown in Table \ref{tab:synthetic}.

\begin{table*}[t!]
\centering
\begin{tabular}{lcll}
\Xhline{2\arrayrulewidth}
\textbf{Task} & \textbf{CoT} & \textbf{Prompt} & \textbf{Target}\\
\hline
\textbf{Addition} & \ding{55} & 1463456 + 2107 & 1465563 \\
\textbf{Subtraction} & \ding{55} & 2348233 - 483579 & 1864654 \\
\textbf{Multiplication} \\
\hspace{2ex}$nd \times 1d$ & \ding{55}& 593295 * 7  &  4153065\\
\hspace{2ex}$nd \times md$ & \ding{51}  & 24 * 79 & 24 * (70 + 9) \\ &&& = 24 * 70 + 24 * 9 = 1680 + 216 = 1896 \\
\textbf{Division}\\
\hspace{2ex}$nd \div 1d$ & \ding{55}& 3651803 / 7 & 521686 R 1\\
\hspace{2ex}$nd \div md$ & \ding{51}  & 2551 / 38 & 
2546 - 38 * 60 = 2546 - 2280 = 266 \\ &&&
266 - 38 * 7 = 266 - 266 = 0 \\ &&&
Therefore, 2551 / 38 = 67 \\
\Xhline{2\arrayrulewidth}
\end{tabular}
\caption{\label{tab:synthetic}
Examples of input and output for training and testing in the simplified synthetic environment, which is used for testing the learnability of sub-tasks and ablation studies. Specifically, ``+'', ``-'', ``*'', and ``\textbackslash'' are used for addition, subtraction, multiplication, and division, respectively. Space is inserted between numbers and symbols. The input and output are formatted to mitigate the influence of natural language.
}
\end{table*}

\section{Special Cases}
\label{sec:special_cases}
In general, multi-digit multiplication and division are considered unlearnable, and we use the decomposition method to solve them. However, some special cases within multi-digit multiplication and division are learnable, and in these cases, we omit CoT and generate the direct answer:
\begin{itemize}
    \item For multiplication, one of the two numbers contains only one non-zero digit, such as $857483 \times 400 = 342993200$. This type of task is similar to learnable $n$-digit by 1-digit multiplication, with the zeros being copied at the end of the product.
    \item The dividend is equal to the divisor. In that case, the quotient is identically one. For example, $358 \div 358 = 1$.
    \item The dividend is less than the divisor. In that case, the quotient is zero and the remainder equals the dividend. For example, $423 \div 968 = 0 \text{ R } 423$.
\end{itemize}

\section{Dataset}
\label{sec:dataset}
In general, it is difficult to determine the optimal proportion for each task. The number and composition of data samples also depend on the problem settings (see Section \ref{sec:settings}). We empirically find that $n$-digit by 1-digit multiplication and division may be easier than other tasks, as it requires fewer samples to reach the same level of accuracy as other tasks during task-specific fine-tuning in the simplified synthetic environment. It is noteworthy that the data samples are all randomly generated, so the probability of the occurrence of duplicated samples is very low for large numbers. Therefore, the training loss can reflect the test accuracy on unseen the test set, if the dataset is only trained for one epoch. 
\begin{figure}[t!]
  \centering
  \includegraphics[width=0.5\textwidth]{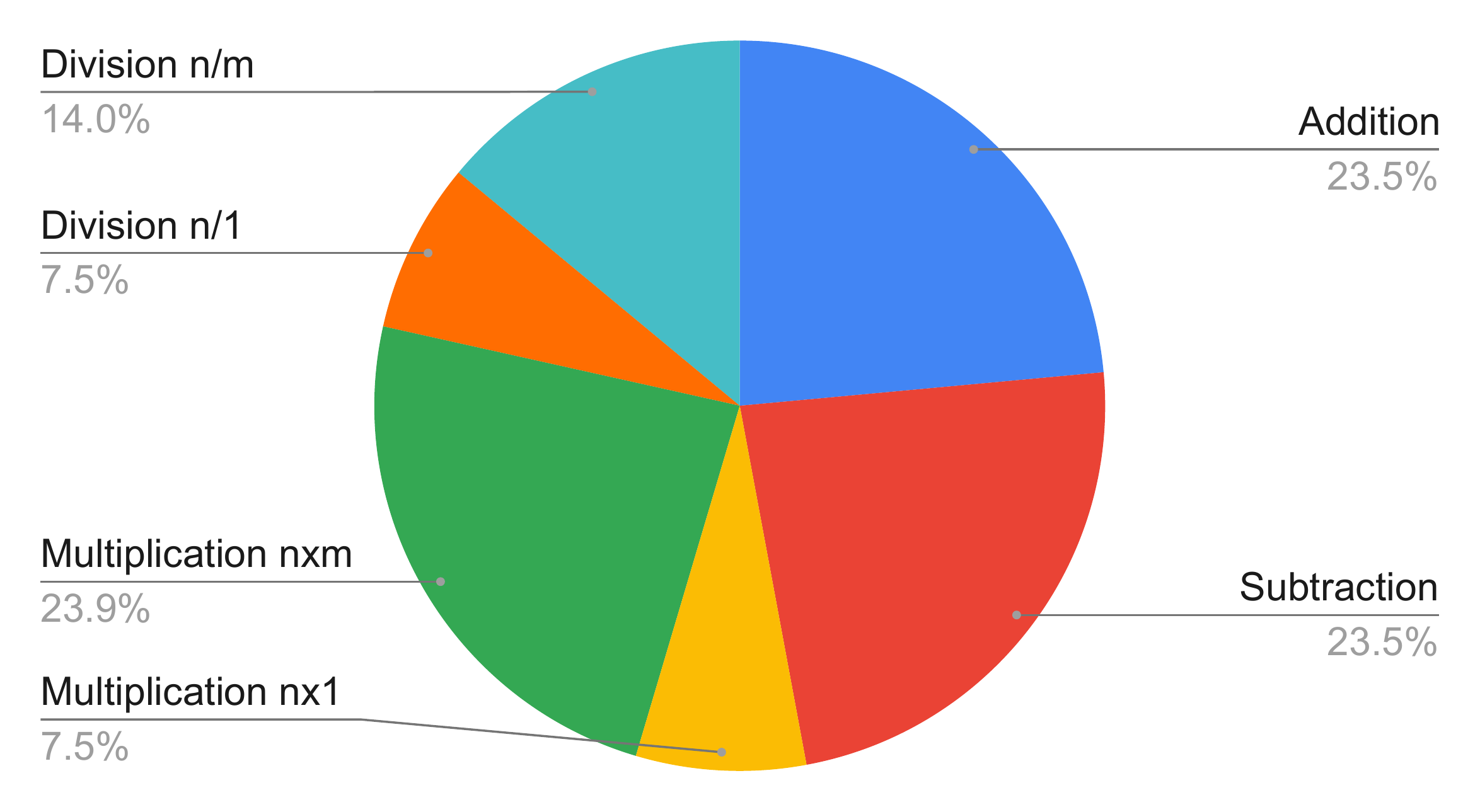}
  \caption{Composition of tasks in the dataset.}
  \label{fig:example2}
\end{figure}
Since the synthetic dataset can be generated very easily, we first create a dataset that contains a sufficient number of data samples for training and then observe the training loss and apply early stopping. We observe that the training loss does not show any significant decrease after training on about one million samples. It should be noted that convergence also depends on other hyper-parameters such as batch size and learning rate. Hence, it is recommended to use a dataset larger than what is necessary and terminate the training process when the training loss no longer decreases.

\section{Ablation Study}
\label{sec:ablation_appendix}
We name the steps (shown in the box below) as (1) extraction, (2) split, (3) expansion, (4) product, and (5, 6, \ldots) adding term by term. The ablation study is performed by removing one particular step while keeping other steps unchanged. We exclude the (1) ``extraction'' and (4) ``product'' steps from the ablation study as it is crucial for multi-digit multiplication.

\begin{tcolorbox}[fonttitle=\bfseries,title=Multiplication,leftlower=0cm, size = normal]
\textbf{Calculate 397 x 4429} \textbackslash nAnswer:
\tcblower
\setlength{\abovedisplayskip}{0pt}
\setlength{\belowdisplayskip}{2pt}
\begin{align}
& 397 \times 4429  \\
&= 4429 \times (300 + 90 + 7)\\
&= 4429 \times 300 + 4429 \times 90 + 4429 \times 7  \\
&= 1328700 + 398610 + 31003\\
& = 1727310 + 31003 \\
& = 1758313
\end{align}
\end{tcolorbox}

For division, the ablation study is performed by removing the middle step (bold) that computes the product for all iterations, while keeping other steps unchanged.
\begin{tcolorbox}[fonttitle=\bfseries,title=Division,leftlower=0cm, size = normal]
\textbf{What is 8914/64?} \textbackslash nAnswer:
\tcblower
\setlength{\abovedisplayskip}{3pt}
\setlength{\belowdisplayskip}{0pt}
\begin{align*}
&8914 - 64 \times 100 = \mathbf{8914 - 6400 = } 2514\\
&2514 - 64 \times 30 = \mathbf{2514 - 1920 = } 594\\
&594 - 64 \times 9 = \mathbf{594 - 576 = } 18\\
&\text{Therefore, } 8914 \div 64 = 139 \text{ R } 18
\end{align*}
\end{tcolorbox}

\newpage
\section{Few-shot Prompting with GPT-4}
\label{sec:fewshot}
Here are two examples showing 3-shot prompting using GPT-4.

\begin{tcolorbox}[fonttitle=\bfseries,title=3-Shot prompting for multiplication,leftlower=0cm, size = normal]
\textbf{Prompt:}

Here are 3 examples of question-solution pairs for solving arithmetic multiplication problems. Solve the last question using the same format given in the solutions.\\

``question'': ``235658 * 91204 = '',\\
``solution'': ``235658 * 91204 = 235658 * (90000 + 1000 + 200 + 4) = 235658 * 90000 + 235658 * 1000 + 235658 * 200 + 235658 * 4 = 21209220000 + 235658000 + 47131600 + 942632 = 21444878000 + 47131600 + 942632 = 21492009600 + 942632 = 21492952232''\\

``question'': ``59909636 * 6070 = '',\\
``solution'': ``59909636 * 6070 = 59909636 * (6000 + 70) = 59909636 * 6000 + 59909636 * 70 = 359457816000 + 4193674520 = 363651490520''\\

``question'': ``7661 * 9461 = '',\\
``solution'': ``7661 * 9461 = 9461 * (7000 + 600 + 60 + 1) = 9461 * 7000 + 9461 * 600 + 9461 * 60 + 9461 * 1 = 66227000 + 5676600 + 567660 + 9461 = 71903600 + 567660 + 9461 = 72471260 + 9461 = 72480721''\\

``question'': ``58061 * 3569 = '',

\tcblower
\textbf{GPT-4:}

To solve the multiplication problem of 58061 multiplied by 3569 in the same format, we'll distribute 58061 across the sum of the digits of 3569 multiplied by their respective places.

So,

``solution'': ``58061 * 3569 = 58061 * (3000 + 500 + 60 + 9) = 58061 * 3000 + 58061 * 500 + 58061 * 60 + 58061 * 9 = 174183000 + 29030500 + 3483660 + 522549 = 203213500 + 3483660 + 522549 = 206697160 + 522549 = 207219709''
\end{tcolorbox}

\begin{tcolorbox}[fonttitle=\bfseries,title=3-Shot prompting for division,leftlower=0cm, size = normal]
\textbf{Prompt:}

Here are 3 examples of question-solution pairs for solving arithmetic division problem. Solve the last question using the same format given in the solutions.

``question'': ``440331 / 22 = '',\\
``solution'': ``440331 - 22 * 20000 = 440331 - 440000 = 331\textbackslash n331 - 22 * 10 = 331 - 220 = 111\textbackslash n111 - 22 * 5 = 111 - 110 = 1\textbackslash nTherefore, 440331 / 22 = 20015 R 1''\\

``question'': ``115247727 / 970051 = '',\\
``solution'': ``115247727 - 970051 * 100 = 115247727 - 97005100 = 18242627\textbackslash n18242627 - 970051 * 10 = 18242627 - 9700510 = 8542117\textbackslash n8542117 - 970051 * 8 = 8542117 - 7760408 = 781709\textbackslash nTherefore, 115247727 / 970051 = 118 R 781709''\\

``question'': ``4144442 / 234443 = '',\\
``solution'': ``4144442 - 234443 * 10 = 4144442 - 2344430 = 1800012\textbackslash n1800012 - 234443 * 7 = 1800012 - 1641101 = 158911\textbackslash nTherefore, 4144442 / 234443 = 17 R 158911''\\

``question'': ``262122 / 591 = ''

\tcblower
\textbf{GPT-4:}

To solve the division problem of 262122 divided by 591 in the same format, we'll subtract multiples of 591 from 262122 until we reach a number less than 591, then that's our remainder.

So,

``solution'': ``262122 - 591 * 400 = 262122 - 236400 = 25722\textbackslash n25722 - 591 * 40 = 25722 - 23640 = 2082\textbackslash n2082 - 591 * 3 = 2082 - 1773 = 309\textbackslash nTherefore, 262122 / 591 = 443 R 309''
\end{tcolorbox}

\end{document}